# An Improved Feature Descriptor for Recognition of Handwritten *Bangla* Alphabet

**Nibaran Das, Subhadip Basu, Ram Sarkar, Mahantapas Kundu, Mita Nasipuri, Dipak kumar Basu**

Computer Sc. & Engg. Dept., Jadavpur University,
Kolkata-700032,
India

*Abstract* - Appropriate feature set for representation of pattern classes is one of the most important aspects of handwritten character recognition. The effectiveness of features depends on the discriminating power of the features chosen to represent patterns of different classes. However, discriminatory features are not easily measurable. Investigative experimentation is necessary for identifying discriminatory features. In the present work we have identified a new variation of feature set which significantly outperforms on handwritten Bangla alphabet from the previously used feature set. 132 number of features in all viz. modified shadow features, octant and centroid features, distance based features, quad tree based longest run features are used here. Using this feature set the recognition performance increases sharply from the 75.05% observed in our previous work [7], to 85.40% on 50 character classes with MLP based classifier on the same dataset.

*Key words*: Handwritten Bangla alphabet, discriminating features, quad tree based longest run, MLP

## 1.0 INTRODUCTION

Any object or pattern, which can be recognized and classified, possesses a number of discriminatory properties of features. More preciously, features are the descriptive measures of a pattern which characterize the membership of a pattern in a certain class. The task of feature extraction is to reduce the data by measuring "features" or "properties" that distinguish between different characters. These features are then passed to a classifier that evaluates the evidence presented and makes a final decision. It is evident that the investigation of feature extraction methods has gained considerable attention since a discriminative feature set is considered the most important factor in achieving high recognition performance. Trier et al [1] present an interesting survey of feature extraction methods for off-line recognition of segmented characters. In the same direction, an interesting review of shape analysis techniques is presented by Loncaric [2]. However, discriminatory features are not easily measurable. Investigative experimentation is necessary for identifying discriminatory features before the design of pattern classifier.

Among Indian scripts, Bangla has started to receive attention for OCR related research in the recent years although evidences of research are few [3,4,5,6,7,8]. The complex Nature of Bangla alphabet coupled with numerous variations of writing styles of individuals might be the reason behind it. Compared to Roman script, basic Bangla character-set consists of a much larger number of characters. If we analyze the Bangla script we have found there are 50 basic characters. Though there also exists 10 modifiers and around 260 compound characters in said script but our research interest, for the time being, is limited within basic characters only. Among these basic characters, 11 are vowels and 39 are consonants. Fig 1. Shows the typical handwritten basic character patterns of Bangla script. The writing style of Bangla script is horizontal and left to right and the script is not case-sensitive. And some characters therein resemble pair wise so closely that the only sign of small difference left between them is a period or a small line.

Two of the important research contributions relating to OCR of Bangla characters involve a multistage approach developed by Rahman et al. [3] and an MLP classifier developed by Bhowmik et al. [4]. The major features used for the multistage approach include Matra, upper part of the character, disjoint section of the character, vertical line and double vertical line. And, for the MLP classifier, the feature set is constructed from the stroke feature of characters. In the work of Bhowmik et al[4], a large size data set of 25,000 samples, collected from different sections of population, is used for testing performances of the MLP classifier. The size of the input feature vector chosen for the work is 200.

K. Roy et al.[8] developed a quadratic classifier based approach where features are calculated from the directional chain code histogram for Bangla basic characters recognition. In the work of Bhattacharya et al.[5], a two-stage approach is adopted to classify 50 handwritten Basic characters and 10 numeric digits of Bangla script. In this approach also a coarse or a group based coarse classification of an unknown pattern in first stage is followed by a finer classification in the second stage. Based on the similarity of shapes, 57 pattern classes are identified for final classification. These pattern classes are clustered into 11 groups for coarse classification. In another work, Bhattacharya et al.[6] proposed a similar two stage approach for recognition of 50 Basic characters of handwritten Bangla script. Chain-code histogram features are used in both the cases for classification through MLP based classifiers

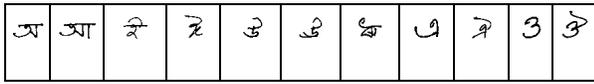

(a)

(b)

**Fig 1. Sample images of 50 handwritten characters of *Bangla* digit and alphabet.**
    **(b). Vowels of *Bangla* script**
    **(c). Consonants of *Bangla* script**

Oh et.al. [10] proposed a new approach for combination of multiple features in handwriting recognition problem. In the said scheme two different sets of feature vectors are designed, where one feature-set is designed to be used by all the classes and in the other set, class-dependent feature sets are designed for patterns of each class. A combination algorithm is finally developed to combine the said feature sets for classification of patterns using a neural network classifier.

S. Basu et al. [9] developed a 76 dimension feature set, for recognition of handwritten Bangla alphabet using an MLP based classifier. The success rate achieved in the said work is 75.05. In one of the recent works S. Basu et al.[7] developed a two stage hierarchical approach for recognition of handwritten Bangla characters. Using the 76 topological features, as described in [9], in two stages the overall recognition accuracy of 80.58% is reported on a reduced 36 class data set. In the present work, we have introduced some new feature descriptor and used them in combination with some features previously used in [7,9].

This motivate us to develop a new variation of feature set which will represent more appropriately than the previously designed feature set [9] for recognition of the same database of Bangla handwritten characters. Using this improved feature set the recognition performance increases sharply from the 75.05% to 85.40% on 50 classes with MLP based classifier on the same dataset.

**2.0 THE FEATURE SET**

The feature set selected for the present work consists of 132 features, which include 24 shadow features, 16 centroid features, 8 distance based features and 84 quad tree based longest run features. The features are computed from 64x64 pixel size binary images of alphabetic characters of Bangla script..

*2.1. Shadow Features*
For computing shadow features [9], each character image is enclosed within a minimal square, divided into eight octants. Lengths of projections of character images on three sides of each octant are then computed. Finally lengths of all such projections on each of the 24 sides of all octants are summed up to produce 24 shadow features of the character image under consideration. For taking the projection of an image segment on one side of an octant, existence of a fictitious light source in the opposite side is assume.

*2.2. Centriod Features*
Coordinates of centroids of black pixels[] in all the 8 octants of a character image are considered to add 16 features in all to the feature set.

*2.3 Distance based features:*
To compute the distance-based features we have partitioned the character images in four quadrants. For every quadrant, maximum horizontal and diagonal distances from image boundary to charater boundary have been calculated. How these features have been calculated is shown in Fig 2. Thus, for four directions 2x4=8 features have been calculated in all.

*2.4. Quad-tree based Longest run features*

*2.4.1 Longest-run Features*
Within a rectangular image region of a character, a longest run feature [4] are computed in four directions, viz, row wise, column wise and along the directions of two major diagonals. The row wise longest run feature is computed by considering the sum of the lengths of the longest bars that fit consecutive black pixels along each of all the rows of the region.

In fitting a bar with a number of consecutive black pixels within a rectangular region, the bar may extend beyond the In fitting a bar with a number of consecutive black pixels within a rectangular region, the bar may extend beyond the boundary of the region if the chain of black pixels is continued there. The three other longest-run features within the rectangle are computed in the same way. Each of the longest run feature values is to be normalized by dividing it with the product of the height (h) and the width (w) of the entire image. The product, h x w, represents the sum of the lengths of the bars that fit consecutive black pixels individually in each of the four directions within the region completely filled with black pixels.

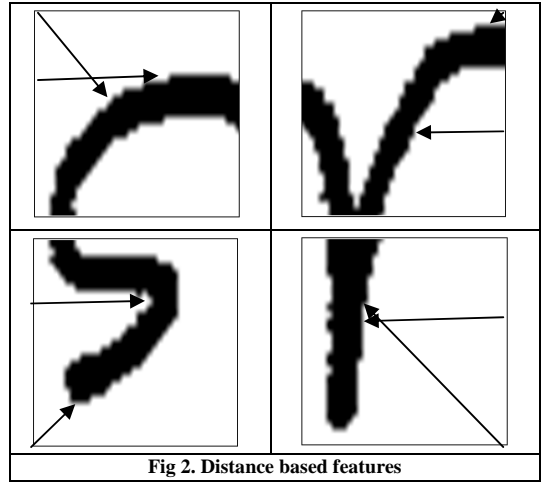

**Fig 2. Distance based features**

*2.4.2. Quad-tree Structure*
A quad-tree is a tree data structure in which each node except the leaf nodes has up to four children. Quad-trees are most often used for representation of a two dimensional space by recursively subdividing it into four equal quadrants or regions. In the current work, we have used a modified version of quad tree structure to partition any digit pattern into multiple sub-images. Here, partitioning a character pattern (or a subpart of it) into 4 regions is done by drawing a horizontal and a vertical line through the Centre of Gravity (CG) of black pixels in that region. If the depth of the quad-tree structure is d, then total number of sub images for each digit pattern at leaf nodes would be 4d. The coordinates of the CG of any image frame, $(C_x, C_y)$, is calculated as follows:

$$C_x = \frac{1}{mn} \sum_{mn} x.f(x,y) \quad ; \quad C_y = \frac{1}{mn} \sum_{mn} y.f(x,y)$$

$$f(x,y) = \begin{cases} 1; & \text{for all black pixels} \\ 0; & \text{otherwise} \end{cases}$$

where, x and y are the coordinates of each pixel in the image of size m x n pixels. Fig. 3 (a) shows sample images, Fig. 5(b) shows equal partitioning and Fig 3(c) shows the CG based partitioning for generating the quad-tree structure of depth 2 for each of the sample images. For each sub image at any node of the quad-tree structure, 4 longest-run features are computed. Partitioning any character pattern using CG based quad tree structure is a novelty of the current work. Equal partitioning, as usually done in many approaches, often generates less informative sub-images in comparison to the CG based partitioning. A sample digit image with equal partitioning structure is shown in Fig.3(c). Comparing Figs. 3(b) and 3(c), it may be observed that the equal partitioning structure generates many sub-images with no information, which is avoided in the current CG based quad-tree structure.

In the current work, we have considered the depth of the quad-tree structure (d) as 2 which consists of a root node, 4 nodes at depth 1 and 42 nodes at depth 2. Thus, the total number of nodes in the quad tree structure is 21(=1+4+16). Altogether 84(=21x4) longest run features are computed for each character image.

Thus total 132 features(=24+16+8+84) are considered for the present work.

## 3.0 The MLP Classifier

In the present work, an MLP classifier [9] is employed for recognition of unknown Bangla basic characters using the

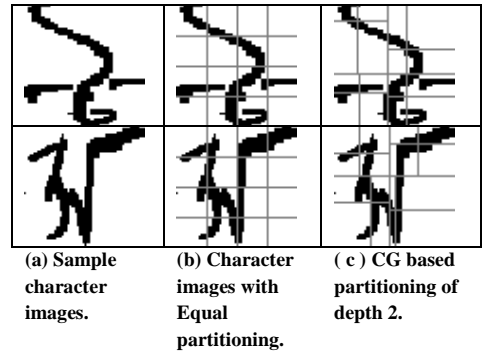

(a) Sample character images.   (b) Character images with Equal partitioning.   (c) CG based partitioning of depth 2.

**Fig. 3. Different image partitioning schemes for different samples**

above mentioned 132 feature set. MLP has been chosen because it is a kind of feed forward Artificial Neural Networks (ANNs) in general famous for their learning and generalization abilities, necessary for dealing with imprecision in input patterns.

## 4.0 RESULTS AND DISCUSSION

For preparation of the training and the test sets of samples, a database of 10,000 alphabetic character samples is formed by collecting optically scanned handwritten characters specimens of 50 alphabetic symbols from each of 200 people of different age groups and sexes. A training set of 8000 samples and a test set of 2000 samples are then formed through random selection of character samples of each class from the initial database in equal numbers. All these samples are scaled to 64x64 pixel images first and then converted to binary images through thresholding.

For the present work, a single layer MLP, i.e., an MLP with one hidden layer is chosen. This is mainly to keep the computational requirement of the same low without affecting its function approximation capability. According to Universal Approximation theorem [5], a single hidden layer is sufficient to compute a uniform approximation to a given training set.

To design an MLP for classification of handwritten alphabetic characters, several runs of BP algorithm with learning rate ($\eta$) = 0.8 and momentum term ($\alpha$)=0.7 are executed for different numbers of neurons in its hidden layer. Recognition performances of the MLP on the training and the test sets observed from this experimentation are given in Table1.

Curves showing variation of the Recognition performance of the MLP, for both the test and the training sets, with increase in the number of neurons in its hidden layer are plotted in Fig. 6 from the Table 1. It is required to fix up the number of neurons in the hidden layer of MLP so that it can show the optimal recognition performance on the test set. Recognition performances of the MLP, as observed from the curve shown in Fig 4., initially rise as the number of neurons in the hidden layer is increased and falls after the same crosses some limiting value. It reflects the fact that for some fixed training and test sets, learning and generalization abilities of the MLP improve as the number of neurons in its hidden layer as increases up to certain limiting value and any further increase in the number of neurons in the hidden layer thereafter degrades the abilities. It is called the over-fitting problem.

| Table 1 | | | | | | | | | | |
|---|---|---|---|---|---|---|---|---|---|---|
| *No of Hidden neurons* | 40 | 45 | 50 | 55 | 60 | 65 | 70 | 75 | 80 | 85 |
| **Percentage recognition rate on test samples** | 83.05 | 83.05 | 83.10 | 83.15 | 84.60 | 85.40 | 84.20 | 84.25 | 85.00 | 84.45 |

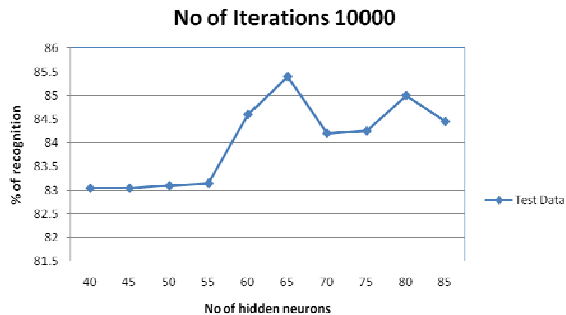

**Figure 4. Curves show variation of recognition performances of the MLP as the number of neurons in its hidden layer is increased for the test sets of Numerals**

The optimal recognition performance of the MLP is observed at a point, on the curve of Fig.6, where the number of neurons in its hidden layer is set to 65. Thus the number of neurons in the hidden layer of the MLP is finally fixed up to 65. With this, the design process is completed producing an MLP (132-65-50) for recognition of handwritten alphabet on the basis of the feature set explained before. This will show the enhancement of previous result 75.05%. It is due to the discriminate property of newly designed feature set more appropriate for Bangla alphabet. Some samples of misclassified character images are shown in Fig. 7(a-d).

In this work we emphasizes on finding an effective feature descriptor for the Bangla alphabet. Recognition performances of the MLP can be further improved firstly by adding newer variation of handwritten alphabetic samples to the training set and secondly using two pass approaches. Where in the first pass mutually misclassified samples can be considered as a group and in the second pass they may be recognized using more selective features [9] and refined the earlier decision by combining the local and the global features. The work presented here can have useful application in the development of a complete OCR system for handwritten Bangla text.

## ACKNOWLEDGEMENT

Authors are thankful to the "Center for Microprocessor Application for Training Education and Research", "Project on Storage Retrieval and Understanding of Video for Multimedia" and Computer Science & Engineering Department, Jadavpur University, for providing infrastructure facilities during progress of the work.

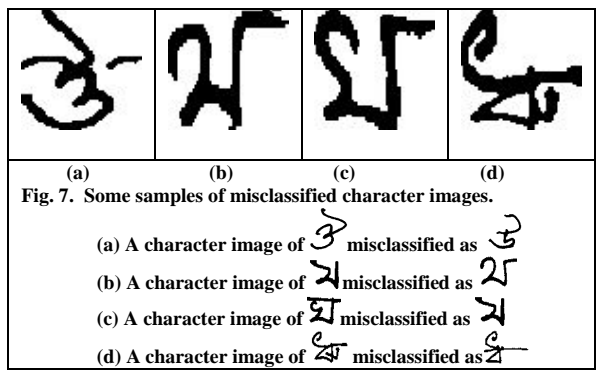

(a) (b) (c) (d)
**Fig. 7. Some samples of misclassified character images.**
(a) A character image of ঐ misclassified as ঔ
(b) A character image of ম misclassified as য
(c) A character image of ঘ misclassified as য
(d) A character image of ক্ষ misclassified as হ